\pdfoutput=1

\documentclass[11pt]{article}

\usepackage[final]{acl}

\usepackage{times}
\usepackage{latexsym}

\usepackage[T1]{fontenc}

\usepackage[utf8]{inputenc}

\usepackage{microtype}

\usepackage{inconsolata}
\usepackage{amsmath}
\usepackage{multirow}
\usepackage{multicol}

\usepackage{algorithm}
\usepackage[noend]{algpseudocode}

\usepackage{graphicx}
\usepackage{caption}
\usepackage{subcaption}

\usepackage{xurl} 
\usepackage{breakurl} 
\usepackage{hyperref} 

\usepackage{array}      
\usepackage{booktabs}   

\usepackage[autostyle]{csquotes}
\usepackage{enumitem}

\usepackage{titlesec}
\title{LLM Bias Detection and Mitigation through the Lens of Desired Distributions}

\author{Ingroj Shrestha \\
  University of Iowa \\
  \texttt{ingroj-shrestha@uiowa.edu} \\\And
  Padmini Srinivasan \\
  University of Iowa \\
  \texttt{padmini-srinivasan@uiowa.edu} \\}

\newcolumntype{H}{>{\setbox0=\hbox\bgroup}c<{\egroup}@{}}

\newcommand{\DPF}{$\text{DP}_{\text{female}}$\xspace}
\newcommand{\DPM}{$\text{DP}_{\text{male}}$\xspace}
\newcommand{\DPB}{$\text{DP}_{\text{balanced}}$\xspace}

\usepackage{makecell} 
\usepackage{xspace}
\usepackage{multirow}
\usepackage{array}

\usepackage{enumitem}

\begin{document}
\maketitle

\begin{abstract}

Although prior work on bias mitigation has focused on promoting social equality and demographic parity, less attention has been given to aligning LLM's outputs to desired distributions. For example, we might want to align a model with real-world distributions to support factual grounding. Thus, we define bias as deviation from a desired distribution, which may be an equal or real-world distribution, depending on application goals.
We propose a weighted adaptive loss\footnote{Our code and data are available at \url{https://github.com/IngrojShrestha/bias_through_lens_of_desired_distributions}.}
based fine-tuning method that aligns LLM’s gender–profession output distribution with the desired distribution, while preserving language modeling capability. 
Using 3 profession sets---male-dominated, female-dominated, and gender-balanced---derived from U.S. labor statistics (2024), we assess both our adaptive method for reflecting reality and a non-adaptive variant for equality. 
Across three masked language models, bias is observed under both distributions.  We achieve near-complete mitigation under equality and 30–75\% reduction under real-world settings. Autoregressive LLMs show no bias under equality but notable
bias under real-world settings, with the Llama Instruct models (3.2-3B, 3.1-8B) achieving a 50–62\% reduction.

\end{abstract}

\section{Introduction}
\noindent Large Language Models (LLMs) have achieved remarkable performance across a range of natural language processing (NLP) tasks. However, this success is tempered by the presence of social and representational biases \cite{gallegos-etal-2024-bias,guo-etal-2022-auto,kaneko2022unmasking,nadeem-etal-2021-stereoset}. 
The computer science (CS) literature on LLMs bias typically considers any differences in association between attribute values (e.g., male and female for gender attribute) in a given context (e.g., profession) as an indication of bias. Psychology also views such differences as bias. However, when a model reflects the real world, CS still sees this as bias, whereas psychology considers it an accurate reflection of reality.
Thus, there are two bias viewpoints. Bias is (1) any deviation from an equal (50-50) distribution---often captured by fairness notions such as demographic parity, equalized odds, or equal opportunity \cite{gallegos-etal-2024-bias,
mehrabi2021survey}--- regardless of real-world distributions (2) deviation from real-world distributions.
Less attention has been given in the CS bias literature to this second viewpoint, important in certain applications, such as healthcare, where genetic and biological predispositions (e.g., age, gender) make equality undesirable for LLMs in contexts like health chatbots, and precision medicine.
The problem may be generalized as one where the aim is to align LLM to a user-specified distribution. We address the two specific cases of equal versus real-world distributions. It is of course possible for the two to be the same.
In both cases, fairness can be achieved by adjusting the model's distribution to the desired distribution, which we achieve by fine-tuning.

Fairness with the first bias viewpoint has been explored extensively in CS \cite{gallegos-etal-2024-bias,delobelle2022measuring,guo-etal-2022-auto,stanczak2021survey}. 
Prior work typically measures bias at a granular level using attribute–target combinations, often with or without templates, where gendered attribute words (e.g., male- and female-associated terms) are paired with target concepts (e.g., professions) to compute association scores that quantify the strength of gender–profession associations. These associations are estimated using either template-based probe sentences or corpus-based contextual embeddings \cite{shi2024general,yang2023adept,guo-etal-2022-auto,limisiewicz-marecek-2022-dont,garimella-etal-2021-intelligent}, and evaluated on out-of-distribution bias benchmarks such as WinoBias \cite{zhao-etal-2018-gender} and Winogender \cite{rudinger-etal-2018-gender} to test generalization beyond the training templates.
In contrast, our work shifts focus to a coarser, distributional perspective: we analyze the gender distribution across professions, for example, for \textit{dental assistant} it is: 8\% male, 92\% female, and aim to align these distributions to assess and mitigate bias.

Fairness with the second bias viewpoint requires more attention. When models do not reflect reality ---whether factual or perceptual--- they are likely to generate misleading information and hallucinations it is critically important to address such hallucinations \cite{sahoo-etal-2024-comprehensive,niu-etal-2024-ragtruth,su-etal-2024-unsupervised}.
Aligning LLMs with real-world trends also enhances fairness in fact-checking, thus, increasing model reliability and trustworthiness. Recent techniques such as Retrieval Augmentation Retrieval (RAG) \cite{lewis2020retrieval} and Reinforcement Learning with Human Feedback (RLHF) \cite{ouyang2022training,ziegler2019fine} are designed to guide models toward producing outputs grounded in factual information and real-world context. Our second view of bias and fairness aligns with these. 

Based on the selected bias viewpoint, we propose a bias mitigation strategy that fine-tunes the LLMs using a tailored loss function to recalibrate their output distribution towards a desired target. While existing methods often focus on reducing task-specific disparities, less attention is given to the foundational prediction behavior of pre-trained LLMs.
In contrast, we align LLM's output distribution with a desired target distribution during fine-tuning while preserving performance---measured by \textit{MLM loss} and downstream GLUE evaluation for masked language models (MLMs) and by \textit{perplexity} and the LM Evaluation Harness on 5 benchmarks for auto-regressive language models (ALMs). To promote balanced and stable learning across profession groups (male-dominated, female-dominated, balanced), we introduce a weighted adaptive KL loss that dynamically adjusts updates based on group specific dynamics.

\noindent
\textbf{Contributions of our research:} 

\begin{enumerate}[noitemsep,nolistsep,topsep=0pt,leftmargin=*]
    \item In addition to debiasing towards \textit{equality}, we also define debias by aligning a deviation from a \textit{desired distribution}, even if this reflects inequality across groups. For equality, unlike prior fine-grained methods, our approach mitigates bias at a coarser-grained level across groups.

    \item We propose a weighted adaptive loss-based approach to mitigate bias by aligning LLM's output distribution with a desired distribution, 
    while preserving language modeling capability.
\end{enumerate}

\section{Method}

We follow the standard template-based approach to estimate bias in MLMs \cite{gallegos-etal-2024-bias,cimitan-etal-2024-curation,nozza-etal-2022-pipelines} and extend this approach to ALMs.
A template refers to a sentence structure that includes \textit{attribute} (a demographic group against which bias is studied), \textit{target} (the context or domain in which the bias is analyzed), and other neutral words. 
We focus on a single attribute-target pair: gender (attribute) and profession (target).
We chose profession as the target due to the availability of real-world gender distributions to ground our bias analysis.

We use six templates (Appendix \ref{templates} Table \ref{tab:gender-bias-in-profession-general}) to analyze bias in relation to gender-profession distribution. The first five templates are adapted from \citet{bartl-etal-2020-unmasking}, and we added the final one. This aligns with the common practice of using 2-5 templates as discussed in \citet{shrestha-etal-2025-robust}.

\noindent \textbf{Attributes:} We used 11 pairs of binary gender-denoting words (Appendix Table \ref{tab:gendered-words}), adapted from \citet{bartl-etal-2020-unmasking}.

\noindent \textbf{Targets ($\mathcal{R}$):} We used 225 profession data from \citet{bls2024occupations} ($\ge$ 50k employed), grouped by female participation: \textit{male-dominated} (\DPM: [0-30\%]), \textit{female-dominated} (\DPF: [70,100\%]), and \textit{balanced} (\DPB: [45,55\%]).

Instead of extreme cutoffs \cite{bartl-etal-2020-unmasking}, we use [0, 30\%] and [70,100\%] to include both strongly and moderately dominated professions. The [45,55\%] range approximates a nearly equal gender distribution, with ±5\% margin to allow natural variations and maintain balanced representation.

We also shortened the profession titles, as in \citet{bartl-etal-2020-unmasking}, to improve compatibility with the LLMs vocabulary by increasing the likelihood that profession terms would appear within it. 
Titles were lowercased, singularized, and simplified profession titles to reflect primary roles (e.g., \textit{Railroad conductors and yardmasters} became \textit{conductor}).

\subsection{LLM's gender-profession distribution}

For MLMs, we followed \citet{shrestha-etal-2025-robust} to measure the gender-profession association score. We mask the attribute (ideally, a gendered word) in the probe sentence derived from the template and compute the likelihood of predicting the original gendered word. 
To account for the possibility that LLM could be overly trained on a particular gender, we also compute a prior by masking both the attribute and target, and estimating the likelihood of the attribute. Association score $S$ is then obtained as the log-likelihood ratio of these two likelihoods. 

\noindent\textbf{Normalized gender-profession distribution:} After computing the association score between each male/female gendered word and profession across templates (Appendix Table \ref{tab:gender-bias-in-profession-general}), we aggregate and normalize these scores such that the resulting gender distribution for each profession sums to 1.

Let $\mathcal{M}$ ($\mathcal{F}$) denote the set of male- (female-) gendered words. For gender $g \in \{\text{male}, \text{female}\}$, we define the corresponding word set as $\mathcal{G}_g = \mathcal{M}$ if $g = \text{male}$, and $\mathcal{G}_g = \mathcal{F}$ otherwise. The variable $a$ denotes a gendered word selected from $\mathcal{G}_g$.

Let $S^{(g)}_{a,r,t}$ represent the association score for gender $g$, gendered word $a$, profession $r$, in template $t$. 
We compute the aggregated score $S_r^{(g)}$ across all templates and words associated with gender $g$.
The normalized gender-probability distribution $p_{\text{pred}}^{(r)}(g)$ is then obtained by applying a softmax over genders for each profession $r \in \mathcal{R}$, as shown in Eq.~\ref{normalized-gender-prof-dist}

\vspace{-0.5em}
{\footnotesize
\begin{equation}
S_r^{(g)} = \sum_{t \in T} \sum_{a \in \mathcal{G}_g} \exp\left(S^{(g)}_{a,r,t}\right)
\label{eq:exponent-of-association-score}
\end{equation}
}

\vspace{-0.5em}
{\footnotesize
\begin{equation}\label{normalized-gender-prof-dist}
p_{\text{pred}}^{(r)}(g) = \frac{S_r^{(g)}}{\sum_{g' \in \{\text{male}, \text{female}\}} S_r^{(g')}}
\end{equation}

}

For ALM, we use sentence loss as a proxy for association score, as in \citet{hossain-etal-2023-misgendered}.
Since higher loss indicates weaker association, we negate the exponent in Eq.\ref{eq:exponent-of-association-score} and use $\exp\left(-S^{(g)}_{a,r,t}\right)$ instead of $\exp\left(S^{(g)}_{a,r,t}\right)$. This ensures that lower loss (stronger association) yields higher gender-profession distribution.

\subsection{Bias detection}\label{bias-detection}

\noindent Bias is quantified using the Kullback–Leibler (KL) divergence between the predicted distribution $p_{\text{pred}}^{(r)}(g)$ and the desired distribution $p_{\text{true}}^{(r)}(g)$.
\( p_{\text{pred}}^{(r)}(g) \) denotes the predicted gender distribution for profession \( r \) obtained from LLM, and \( p_{\text{true}}^{(r)}(g) \) is the corresponding desired distribution.

To detect bias, we compute the KL divergence between predicted and desired gender distributions for each profession $r$, averaging over male and female distributions.
(Eq. \ref{eq:kl_avg}\footnote{
Mathematically, $D_{KL}(P \parallel Q)$ measures how much the predicted distribution $Q$ deviates from the true distribution $P$.
In PyTorch, it quantifies how closely model predictions align with the target, effectively capturing the deviation of LLM outputs from the true distribution during fine-tuning.}).

\vspace{-1.5em}
{\scriptsize
\begin{equation}
D_{\text{KL}}\left(p_{\text{true}}^{(r)} \parallel p_{\text{pred}}^{(r)}\right) = 
\frac{1}{2} \sum_{g \in \{\text{male}, \text{female}\}} 
p_{\text{true}}^{(r)}(g) \log \left( \frac{p_{\text{true}}^{(r)}(g)}{p_{\text{pred}}^{(r)}(g)} \right)
\label{eq:kl_avg}
\end{equation}
}

\noindent \textbf{Bias score:} The final bias score is the average KL divergence across all professions (Eq. \ref{eq:bias_score}).

\vspace{-0.5em}
{\small
\begin{equation}
\text{BiasScore} = \frac{1}{|\mathcal{R}|} \sum_{r=1}^{|\mathcal{R}|} D_{\text{KL}}^{(r)}
\label{eq:bias_score}
\end{equation}
}

where, $D_{\text{KL}}^{(r)}$ represents KL divergence for profession $r$. 
An ideal unbiased model is one with bias score close to zero.

\subsection{Bias mitigation} \label{bias-mitigation-method}

\noindent We fine-tune gender-profession distribution in LLM to a desired distribution using templates, gendered words, and profession. 
This is done separately for two targets: \textit{equal} (50-50) and \textit{real-world} distributions, with one model for each.
We used all three categories --- \DPM,  \DPF and  \DPB for both fine-tuning and evaluation. 

\noindent \textbf{Non-adaptive KL Loss ($\mathcal{L_{\text{KL,uniform}}}$):}
To guide the model towards the desired distribution, we define loss as KL divergence of the LLM predicted gender-profession distribution ($p_{\text{pred}}^{(r)}(g)$) from the desired distribution ($p_{\text{true}}^{(r)}(g)$) across professions. Fine-tuning minimizes this loss.
The overall loss, $\mathcal{L_{\text{KL,uniform}}}$ (Eq. \ref{eq:no-adaptive-loss}), gives equal weight to each profession, regardless of its profession category.

\vspace{-0.5em}
{\small
\begin{equation}
\mathcal{L_{\text{KL,uniform}}} = \frac{1}{|\mathcal{R}|} \sum_{r \in \mathcal{R}} \mathcal{L}^{(r)} = \text{BiasScore}
\label{eq:no-adaptive-loss}
\end{equation}
}

\vspace{-0.5em}
{\small
\begin{equation}
\mathcal{L}^{(r)} = D_{\text{KL}}\left(p_{\text{true}}^{(r)} \,\|\, p_{\text{pred}}^{(r)}\right)
\label{eq:per-professon-loss}
\end{equation}
}

\noindent \textbf{Weighted adaptive KL loss ($\mathcal{L_{\text{KL,weighted\_adaptive}}}$):}
To better balance learning across profession categories, we propose a weighted adaptive loss approach not previously explored in the context of bias mitigation.
Instead of computing a uniform loss (Eq. \ref{eq:no-adaptive-loss}), we make the loss computation profession-group aware. 
This design was motivated by validation-set analysis, where some groups (e.g., in BERT-base: $\textbf{DP}_{\text{male}}:0.232/0.087$, $\text{DP}_{\text{female}}:0.038/0.001$ and $\text{DP}_{\text{balanced}}:0.085/0.007$) showed higher KL means and variances, indicating greater deviation and instability. 
This motivated the use of both adaptive loss and stability-aware weighting.

\textit{Adaptive loss:} During tuning, we group each training batch $\mathcal{B}$ by profession category, $c \in \{\text{DP}_{\text{male}}, \text{DP}_{\text{female}}, \text{DP}_{\text{balanced}} \}$. 
We compute an adaptive loss for a profession category by dividing the current batch's KL divergence loss, $\mathcal{L}_{\text{cur}}^{(c)}$, by the exponentially updated moving average 
(shown in Eq. \ref{eq:EMA-KLMean}, where $\beta$ refers to momentum parameter controlling how much weight is given to the old KL mean versus the current batch KL mean) 
of the KL loss for that profession category, $\mu_{\text{KL,new}}^{(c)}$, as shown in Eq. \ref{eq:adaptive-loss-with-alpha}. 
This KL mean ---updated with each batch--- captures how the model has historically performed on that group, including the current batch. Computing adaptive loss in this way ensures that groups with consistently high KL divergence (i.e., larger deviation from the desired distribution) do not disproportionately dominate the overall loss, thus promoting balanced learning across all groups. 

\vspace{-0.6em}
{\small
\begin{equation}
\hat{\mathcal{L}}_{\text{cur}}^{(c)} = \frac{\mathcal{L}_{\text{cur}}^{(c)}}{\mu_{\text{KL,new}}^{(c)} + \alpha^{(c)}}
\label{eq:adaptive-loss-with-alpha}
\end{equation}
}

\vspace{-0.4em}
{\small
\begin{equation}
\mu_{\text{KL,new}}^{(c)} = \beta \cdot \mu_{\text{KL,old}}^{(c)} + (1 - \beta) \cdot \mathcal{L}_{\text{cur}}^{(c)}
\label{eq:EMA-KLMean}
\end{equation}
}

\textit{Adaptive loss scaling:} To further control update magnitude, we add a small constant $\alpha^{(c)}$ to the denominator (Eq. \ref{eq:adaptive-loss-with-alpha}).
$\alpha$ is set lower (higher) for the profession category with higher (lower) KL divergence from the target distribution, allowing for larger (smaller, more cautious) updates
thus helping regulate how aggressively the model should adapt to each group’s loss. 
Profession category with higher or lower KL are identified using validation set statistics before adjusting the distribution.

\textit{Stability aware weighting to adaptive loss:} After normalization, we apply an adaptive weighting factor $\lambda^{(c)}$ (Eq. \ref{eq:stability-weight}), computed from the variance of KL divergence for group $c$ so far ($\text{Var}^{(c)}$), using Welford’s online algorithm \cite{welford1962note}.
This variance-based weight captures the stability of predictions---groups with higher variance, indicating less stability, are assigned lower weights, leading to slower, more conservative updates. In contrast, lower-variance (more stable) groups are weighted more heavily, enabling faster adaptation. 

\vspace{-0.5em}
{\footnotesize
\begin{equation}
\text{VarFactor}^{(c)} = \frac{1}{1 + \text{Var}^{(c)}}
\label{eq:variace-factor}
\end{equation}
}

\vspace{-1em}
{\tiny
\begin{equation}
\lambda^{(c)} =
\begin{cases}
\max\big( \min( 0.95 \cdot \mu \cdot V,\ 1.5 ),\ 0.8 \big) & \text{if } c \in \text{high-KL group} \\
\max\big( \min( 1.2 \cdot \mu \cdot V,\ 1.5 ),\ 1.0 \big) & \text{otherwise}
\end{cases}
\label{eq:stability-weight}
\end{equation}
}

\vspace{-1em}
{\scriptsize
\[
\mu = \mu_{\text{KL,new}}^{(c)}, \quad V = \text{VarFactor}^{(c)}
\]
}

Overall weighted adaptive loss $\mathcal{L_{\text{KL,weighted\_adaptive}}}$ (Eq. \ref{eq:weighted-adaptive-loss}) ensures that model updates are fair across groups and responsive to each group's learning dynamics. 
Adaptive loss balances weight updates across categories, preventing domination by high-loss groups; adaptive loss scaling controls the magnitude of updates based on initial profession category loss; and stability-aware weighting adjusts update rate based on group stability.
Overall loss is computed by averaging the weighted adaptive losses over all profession-group batches. 
Adaptive weighting is applied only during fine-tuning to guide learning. During evaluation, we compute the bias score using the original KL divergence formulation (Section \ref{bias-detection}), i.e., no adaptive weighting is applied during detection.

\vspace{-2em}
{\small
\begin{equation}
\mathcal{L_{\text{KL,weighted\_adaptive}}} = \frac{1}{|\mathcal{B}|} \sum_{c \in \mathcal{B}} \lambda^{(c)} \cdot \frac{\mathcal{L}_{\text{cur}}^{(c)}}{\mu_{\text{KL,new}}^{(c)} + \alpha^{(c)}}
\label{eq:weighted-adaptive-loss}
\end{equation}
}

\vspace{-1em}

\noindent \textbf{MLM Loss:} We combine KL divergence loss with an MLM loss as a secondary objective to retain masked language modeling ability while adjusting the MLM distribution.
MLM loss is computed on probe sentences—derived from training templates, professions, and gendered words—by masking one token at a time and averaging the likelihood of the original tokens, following \citet{salazar-etal-2020-masked}.
Since most training probes are short (92.5\% $\le$ 8 words; only 7.5\% have 9) we compute the loss across all tokens instead of masking 15\% at random as in \citet{devlin-etal-2019-bert}. The overall objective is:

$\mathcal{L} = \mathcal{L}_{\text{KL}} + \gamma  \cdot \mathcal{L}_{\text{MLM}}$

where $\gamma$ is a hyperparameter that controls the relative importance of the MLM loss. Since KL divergence is our primary loss for bias mitigation, we control only the MLM loss via $\gamma$, using $\gamma \in \{0.001, 0.01, 0.1, 0.2, 0.5, 0.8, 1.0 \}$

Unlike MLMs, where a separate MLM loss is added as a secondary objective to preserve language modeling ability during fine-tuning, ALMs inherently optimize for next-token prediction given prior context. In our setup, we use sentence loss as a proxy for association score, which already reflects the model's perplexity and thus captures its language modeling capability. As a result, there is no need to include a separate objective to preserve language modeling ability when fine-tuning ALMs.

\section{Experiment Design}

\subsection{Models assessed}

We evaluate bias across three MLMs: DistilBERT \cite{Sanh2019DistilBERTAD} and two BERT variants (bert-base-uncased and bert-large-uncased) \cite{devlin-etal-2019-bert}, 
and two families of autoregressive Instruct models:  Llama3 (3.1-8B, 3.2-3B, 3.3-70B) and Qwen2 (2.5-7B, 2.5-72B)\footnote{Llama3 and Qwen2.5 are case-sensitive, but we used case-insensitive inputs during fine-tuning for consistency. This may slightly affect performance relative to its intended usage.}. We assess bias in all models but focus mitigation on Llama3.1-8B-Instruct and Llama3.2-3B-Instruct due to resource limitations. For consistency, we fine-tuned all models using case-insensitive probe sentences. 

For MLMs, we performed full fine-tuning given their smaller sizes (66M–340M parameters). For ALMs, we used parameter-efficient fine-tuning via LoRA (<7B) and QLoRA ($\ge$7B, for memory efficiency at larger scales) given its significantly larger size.
LoRA and QLoRA have been shown to perform well on Llama models \cite{xin-etal-2024-beyond}, achieving competitive results with much lower memory and compute by updating only a small number of low-rank matrices instead of the full model weights. This makes them a practical and effective choice for large-scale ALMs.

\subsection{Dataset split}
We use a 65\%-15\%-20\% stratified training-validation-testing split.

\noindent \textbf{Attributes and Targets:}
We use the same set of 
gendered pairs across training, validation, and testing. However, since we shift profession distributions, we use distinct profession sets for each split (Appendix Table \ref{tab:profession_dist}).

\noindent \textbf{Templates:}
We use the same set of templates for training and validation, while a different set of templates for testing. We use 3 templates for each. We split the templates into training/validation and testing by balancing the selection of common or rare templates. Note that sentences derived from a template with lower pseudo-perplexity are common sentences. We use a cut-off of sentence pseudo-perplexity 15 to categorize templates. 

T1 and T6 are rare, with fewer than half of their sentences having perplexity below 15, indicating less predictable language (Appendix Table \ref{tab:ppl_dist}). In contrast, T2–T5 are more common, with over 70\% of sentences below the threshold, suggesting more natural, fluent patterns. We select T1-T3 for training/validation, and the rest for testing, balancing one rare and two common templates in each split.

\subsection{Language modeling capability evaluation}

To assess whether our bias mitigation impacts language modeling, we evaluated model performance on two external corpora: the GAP Corpus \cite{webster-etal-2018-mind} and WikiText-103 (development and test sets) \cite{DBLP:conf/iclr/MerityX0S17}. For MLMs, we report MLM loss; for ALM, we report sentence perplexity, which reflects next token prediction quality.

To further assess preservation of language modeling capability, we also evaluated on downstream tasks: MLMs on GLUE tasks and ALMs using the LM Evaluation Harness \cite{eval-harness} across 5 benchmarks---HellaSwag, LAMBADA (OpenAI), TruthfulQA (generation), MMLU, and GLUE---covering text generation, question answering, classification, and commonsense inference.
While MLMs were fine-tuned and evaluated with case-insensitive inputs, ALM perplexity was computed case-sensitively, matching the model’s original training setup for fair evaluation.

Model hyperparameters and selection methodology are detailed in Appendix \ref{model-configs}.

\section{Baseline}

Our method introduces a unique debiasing strategy, particularly for \textit{real-world} distributions (Section \ref{bias-mitigation-method}) and adopts a stricter bias mitigation setting in specific contextual scenarios (e.g., profession), where fairness is defined with respect to distributional shifts.
Since prior work instead defines fairness primarily under an \textit{equal} distribution, this difference makes direct comparison under \textit{real-world} distributions not possible. So, we only report baseline comparisons in the \textit{equal} distribution setting, to ensure consistency with prior methods.

\noindent \textbf{AttenD:} \citet{gaci-etal-2022-debiasing} mitigates bias by finetuning the attention heads to equalize attention distribution across demographic word pairs (e.g., gender: he/she) in context. 
We trained AttenD using our training dataset. Three MLMs (DistilBERT, BERT-base, BERT-large) were evaluated with the hyperparameters reported in the paper, considering all attention heads, as their results showed this yields better performance.

\noindent \textbf{Counterfactual Data Substitution (CDS):} \citet{bartl-etal-2020-unmasking} mitigates bias by fine-tuning on a gender-swapped version of the GAP corpus, where gendered words and names are substituted to create balanced training data.

\section{Results}\label{results}

Once we select the best configuration (using seed 42) based on the validation set, we provide the results averaged across five seeds as in \citet{hansen-etal-2024-impact}. We adapt the seed values 42, 52, 62, 72, 82 from \citet{zhou2025assessing}.

\vspace{-0.5em}
{
\setlength{\tabcolsep}{4pt}
\renewcommand{\arraystretch}{1.1}

\begin{table*}[htb!]
  \centering
  \tiny
  \captionsetup{font=footnotesize}
  
    \begin{tabular}{ll|cccc|ccc|cccc}

   \hline

    \multirow{3}[0]{*}{\makecell[l]{\textbf{Desired}\\\textbf{Distribution}}}

    & \multirow{3}[0]{*}{\textbf{Model}}
    
    & \multicolumn{4}{c|}{\textbf{Profession Category}} & \multicolumn{3}{c|}{\textbf{MLM loss}} & \multirow{3}[0]{*}{\textbf{Epoch}} & \multicolumn{1}{c}{\multirow{3}[0]{*}{\textbf{$\mathcal{B}$}}} & \multicolumn{1}{c}{\multirow{3}[0]{*}{\textbf{$\beta$}}} & \multirow{3}[0]{*}{\textbf{$\gamma$}} \\
    
    \cline{3-9}
    {}   &   {}  & \multicolumn{1}{p{3.41em}}{\textbf{$\text{DP}_{\text{male}}$\newline{}(KL)}} 
    & \multicolumn{1}{p{4em}}{\textbf{$\text{DP}_{\text{female}}$\newline{}(KL)}} 
    & \multicolumn{1}{p{4.5em}}{\textbf{$\text{DP}_{\text{balanced}}$\newline{}(KL)}} & \multicolumn{1}{p{2.6em}|}{\textbf{ALL\newline{}(KL)}} 
    & \multicolumn{1}{p{5em}}{\textbf{GAP\newline{}corpus}} 
    & \multicolumn{1}{p{6em}}{\textbf{WikiText-103\newline{}(test)}} & \multicolumn{1}{p{6em}|}{\textbf{WikiText-103\newline{}(dev)}} &  &       &  &  \\
    
    \hline
    
    \multirow{3}[0]{*}{equal} 
    
        & Base Model & 0.020 & 0.045 & 0.029 & 0.032 & 0.477 & 0.419 & 0.423 &       &       &       &  \\
        \cline{2-13}
        & AttenD & 3.7E-4 &	3.8E-4 & 2.6E-4 & 3.6E-4 & 10.8 & 10.8 & 10.8 &       &       &       &  \\

        &  \% drop & 98.1\%$\dag$ & 99.2\%$\dag$ & 99.1\%$\dag$ & 98.9\%$\dag$ & -2166\% & -2490\% & -2466\% &       &   {}  &  {}  &  \\
        
        \cline{2-13}
        
        &  + \textbf{$\mathcal{L_{\text{KL,uniform}}}$} & 8.6E-5 & 1.4E-5 & 1.0E-5 & 1.1E-5 & 0.515 & 0.444 & 0.448 & \multirow{2}[0]{*}{4} & \multirow{2}[0]{*}{5} & \multirow{2}[0]{*}{-} & \multirow{2}[0]{*}{-} \\
        
        &  \% drop & 99.6\%$\dag$ & 99.7\%$\dag$ & 99.6\%$\dag$ & 99.6\%$\dag$ & -8.0\% & -6.0\% & -5.9\% &       &   {}  &  {}  &  \\

    \hline
    \multirow{9}[0]{*}{\begin{tabular}[c]{@{}c@{}}\text{real-}\\\text{world}\end{tabular}} 
    
    & Base Model & 0.189 & 0.066 & 0.028 & 0.107 & 0.477 & 0.419 & 0.423 &       &       &       &  \\
    \cline{2-13}
    
    & + $\mathcal{L_{\text{KL,weighted\_adaptive}}}$ & 0.052 & 0.028 & 0.020 & 0.036 & 0.573 & 0.483 & 0.488 & \multirow{2}[0]{*}{7} & \multirow{2}[0]{*}{5} & \multirow{2}[0]{*}{0.95} & \multirow{2}[0]{*}{-} \\
        
    &  \% drop & 72.4\%$\dag$ & 57.0\%$\dag$ & 27.5\%$\dag$ & 66.4\%$\dag$ & -20.1\% & -15.4\% & -15.5\% &       &       &       &  \\

    \cline{2-13}
    &   \makecell[l]{+ $\mathcal{L_{\text{KL,weighted\_adaptive}}}$ - $\alpha$} & 0.051 & 0.031 & 0.019 & 0.037 & 0.583 & 0.489 & 0.494 & \multirow{2}[0]{*}{8} & \multirow{2}[0]{*}{5} & \multirow{2}[0]{*}{0.95} & \multirow{2}[0]{*}{-} \\
    
    &  \% drop & 72.9\%$\dag$ & 52.4\%$\dag$ & 30.2\%$\dag$ & 65.7\%$\dag$ & -22.2\% & -16.7\% & -16.8\% &       &       &       &  \\
    
    \cline{2-13}
    
    & + $\mathcal{L_{\text{KL,uniform}}}$ & 0.049 & 0.037 & 0.022 & 0.039 & 0.510 & 0.440 & 0.444 & \multirow{2}[0]{*}{8} & \multirow{2}[0]{*}{5} & \multirow{2}[0]{*}{-} & \multirow{2}[0]{*}{-} \\

    & \% drop & 74.2\%$\dag$ & 43.2\%$\dag$ & 20.4\%$\dag$ & 63.9\%$\dag$ & -6.9\% & -5.0\% & -5.1\% &       &       &       &  \\
    
    \cline{2-13}
    
    & \textbf{\makecell[l]{+ $\mathcal{L_{\text{KL,weighted\_adaptive}}}$ + $\mathcal{L}_{\text{MLM}}$}} 
    
    & 0.070 & 0.029 & 0.019 & 0.043 & 0.510 & 0.438 & 0.443 & \multirow{2}[0]{*}{7} & \multirow{2}[0]{*}{5} & \multirow{2}[0]{*}{0.95} & \multirow{2}[0]{*}{0.2} \\

    & \% drop & 63.2\%$\dag$ & 56.4\%$\dag$ & 31.3\%$\dag$ & 59.8\%$\dag$ & -6.9\% & -4.7\% & -4.9\% &       &       &       &  \\

      \hline
    \end{tabular}
    {\captionsetup{skip=2pt}
    \caption{Bias mitigation and language modeling performance: MLM loss, GLUE evaluation (Appendix Table \ref{tab:mlm-glue-results}).
    Results are averaged across five seed runs (\textbf{DistilBERT}).
    \textit{Base Model} refers to pre-trained MLMs. \% drop indicates reduction in bias or MLM loss of fine-tuned model relative to \textit{Base Model}.
    \textbf{Baseline}: \textit{AttenD}.
    $\dag$ indicates a statistically significant bias reduction. ALL: includes professions from all three profession categories.
    $\mathcal{B}$: training batch size. 
    $\beta$: weight is given to the old KL mean versus the current batch KL mean in adaptive loss, $\gamma$: relative importance to MLM loss. 
    Values for epochs, $\mathcal{B}$ and $\gamma$ are those that yielded the best validation dataset performance.}
    
  \label{tab:result-distilbert}
  }
  \vspace{-2.5mm}
\end{table*}
}

\subsection{Results for debiasing MLMs}

{
\setlength{\tabcolsep}{4pt}
\renewcommand{\arraystretch}{1.1}

\begin{table*}[htb!]
    \centering
    \tiny
    \captionsetup{font=footnotesize}
    \begin{tabular}{ll|cccc|ccc|cccc}
   \hline

    \multirow{3}[0]{*}{\makecell[l]{\textbf{Desired}\\\textbf{Distribution}}}

    & \multirow{3}[0]{*}{\textbf{Model}}
    
    & \multicolumn{4}{c|}{\textbf{Profession Category}} & \multicolumn{3}{c|}{\textbf{MLM loss}} & \multirow{3}[0]{*}{\textbf{Epoch}} & \multicolumn{1}{c}{\multirow{3}[0]{*}{\textbf{$\mathcal{B}$}}} & \multicolumn{1}{c}{\multirow{3}[0]{*}{\textbf{$\beta$}}} & \multirow{3}[0]{*}{\textbf{$\gamma$}} \\
    
    \cline{3-9}
    {}   &   {}  & \multicolumn{1}{p{3.41em}}{\textbf{$\text{DP}_{\text{male}}$\newline{}(KL)}} 
    & \multicolumn{1}{p{4em}}{\textbf{$\text{DP}_{\text{female}}$\newline{}(KL)}} 
    & \multicolumn{1}{p{4.5em}}{\textbf{$\text{DP}_{\text{balanced}}$\newline{}(KL)}} & \multicolumn{1}{p{2.6em}|}{\textbf{ALL\newline{}(KL)}} 
    & \multicolumn{1}{p{5em}}{\textbf{GAP\newline{}corpus}} 
    & \multicolumn{1}{p{6em}}{\textbf{WikiText-103\newline{}(test)}} & \multicolumn{1}{p{6em}|}{\textbf{WikiText-103\newline{}(dev)}} &  &       &  &  \\
    
    \hline
    
    \multirow{3}[0]{*}{equal} 
    
        & Base Model & 0.046 & 0.164 & 0.060 & 0.096 & 0.474 & 0.438 & 0.446 &       &       &       &  \\
        \cline{2-13}

        & AttenD & 3.4E-4 &	3.2E-4 & 3.4E-4 & 3.3E-4 & 10.8 & 10.8 	& 10.8 &       &       &       &  \\

        &  \% drop & 99.3\%$\dag$ & 99.8.\%$\dag$ & 99.4\%$\dag$ & 99.7\%$\dag$ & -2176\% & -2368\% & -2325\% &       &   {}  &  {}  &  \\
        
        \cline{2-13}
        
        & \textbf{+ $\mathcal{L_{\text{KL,uniform}}}$} & 5.2E-4 & 3.1E-4 & 6.0E-4 & 4.4E-4 & 0.496 & 0.449 & 0.457 & \multirow{2}[0]{*}{3} & \multirow{2}[0]{*}{5} & \multirow{2}[0]{*}{-} & \multirow{2}[0]{*}{-} \\
        
        &  \% drop & 98.9\%$\dag$ & 99.8\%$\dag$ & 99.0\%$\dag$ & 99.5\%$\dag$ & -4.7\% & -2.4\% & -2.3\% &       &   {}  &  {}  &  \\

    \hline
    \multirow{9}[0]{*}{\begin{tabular}[c]{@{}c@{}}\text{real-}\\\text{world}\end{tabular}} 
    
    & Base Model & 0.270 & 0.040 & 0.059 & 0.136 & 0.474 & 0.438 & 0.446 &       &       &       &  \\
    \cline{2-13}
    
    & + $\mathcal{L_{\text{KL,weighted\_adaptive}}}$ & 0.063 & 0.040 & 0.016 & 0.044 & 0.554 & 0.491 & 0.497 & \multirow{2}[0]{*}{6} & \multirow{2}[0]{*}{5} & \multirow{2}[0]{*}{0.60} & \multirow{2}[0]{*}{-} \\
        
    &  \% drop & 76.7\%$\dag$ & -0.3\% & 73.5\%$\dag$ & 67.3\%$\dag$ & -17.1\% & -12.2\% & -11.5\% &       &       &       &  \\

    \cline{2-13}
    &   \makecell[l]{+ $\mathcal{L_{\text{KL,weighted\_adaptive}}}$ - $\alpha$} & 0.073 & 0.039 & 0.013 & 0.047 & 0.595 & 0.518 & 0.523 & \multirow{2}[0]{*}{8} & \multirow{2}[0]{*}{5} & \multirow{2}[0]{*}{0.60} & \multirow{2}[0]{*}{-} \\
    
    &  \% drop & 73.1\%$\dag$ & 2.4\% & 77.5\%$\dag$ & 65.1\%$\dag$ & -25.6\% & -18.2\% & -17.1\% &       &       &       &  \\
    
    \cline{2-13}
    
    & + $\mathcal{L_{\text{KL,uniform}}}$ & 0.065 & 0.044 & 0.020 & 0.048 & 0.505 & 0.458 & 0.465 & \multirow{2}[0]{*}{8} & \multirow{2}[0]{*}{5} & \multirow{2}[0]{*}{-} & \multirow{2}[0]{*}{-} \\

    & \% drop & 76.0\%$\dag$ & -10.3\% & 65.6\%$\dag$ & 64.9\%$\dag$ & -6.6\% & -4.6\% & -4.2\% &       &       &       &  \\
    
    \cline{2-13}
    
    & \textbf{\makecell[l]{+ $\mathcal{L_{\text{KL,weighted\_adaptive}}}$ + $\mathcal{L}_{\text{MLM}}$}}
    
    & 0.067 & 0.039 & 0.012 & 0.045 & 0.521 & 0.469 & 0.475 & \multirow{2}[0]{*}{6} & \multirow{2}[0]{*}{5} & \multirow{2}[0]{*}{0.60} & \multirow{2}[0]{*}{0.2} \\

    & \% drop & 75.1\%$\dag$ & 2.3\% & 79.6\%$\dag$ & 66.9\%$\dag$ & -10.0\% & -7.1\% & -6.5\% &       &       &       &  \\

      \hline
    \end{tabular}
    {\captionsetup{skip=2pt}
    \caption{Bias mitigation and language modeling performance (\textbf{BERT-base}). See Table \ref{tab:result-distilbert} for cell values and notation details.}
    
  \label{tab:result-bert-base}
  }
\vspace{-6mm}
\end{table*}
}

Tables \ref{tab:result-distilbert} - \ref{tab:result-bert-large} present our results on debiasing MLMs. 
\textit{Base Model} refers to the original model without debiasing, while the rest of the rows represent the effect of using our loss functions to adjust MLM's distribution to a desired distribution. 

For \textit{equal} target distribution, we adjust MLM's distribution to a 50\%-50\% male-female distribution for each profession. So, we only apply non-adaptive loss, with equal weight across professions. For the \textit{real-world} target distribution, where professions vary in gender dominance, we applied a weighted adaptive KL loss to mitigate bias. 

{
\setlength{\tabcolsep}{4pt}
\renewcommand{\arraystretch}{1.1}

\begin{table*}[htb!]
    \centering
    \tiny
    \captionsetup{font=footnotesize}
    \begin{tabular}{ll|cccc|ccc|cccc}
    \hline

    \multirow{3}[0]{*}{\makecell[l]{\textbf{Desired}\\\textbf{Distribution}}}

    & \multirow{3}[0]{*}{\textbf{Model}}
    
    & \multicolumn{4}{c|}{\textbf{Profession Category}} & \multicolumn{3}{c|}{\textbf{MLM loss}} & \multirow{3}[0]{*}{\textbf{Epoch}} & \multicolumn{1}{c}{\multirow{3}[0]{*}{\textbf{$\mathcal{B}$}}} & \multicolumn{1}{c}{\multirow{3}[0]{*}{\textbf{$\beta$}}} & \multirow{3}[0]{*}{\textbf{$\gamma$}} \\
    
    \cline{3-9}
    {}   &   {}  & \multicolumn{1}{p{3.41em}}{\textbf{$\text{DP}_{\text{male}}$\newline{}(KL)}} 
    & \multicolumn{1}{p{4em}}{\textbf{$\text{DP}_{\text{female}}$\newline{}(KL)}} 
    & \multicolumn{1}{p{4.5em}}{\textbf{$\text{DP}_{\text{balanced}}$\newline{}(KL)}} & \multicolumn{1}{p{2.6em}|}{\textbf{ALL\newline{}(KL)}} 
    & \multicolumn{1}{p{5em}}{\textbf{GAP\newline{}corpus}} 
    & \multicolumn{1}{p{6em}}{\textbf{WikiText-103\newline{}(test)}} & \multicolumn{1}{p{6em}|}{\textbf{WikiText-103\newline{}(dev)}} &  &       &  &  \\
    
    \hline
    
    \multirow{3}[0]{*}{equal} 
    
        & Base Model & 0.073 & 0.095 & 0.046 & 0.076 & 0.983 & 0.888 & 0.893 &       &       &       &  \\
        \cline{2-13}
        
        & AttenD & 3.2E-3 & 3.2E-3  & 3.1E-3 & 3.2E-3 & 11.0 & 11.0 & 11.0 &       &       &       &  \\

        &  \% drop & 95.6\%$\dag$ & 96.7\%$\dag$ & 93.2\%$\dag$ & 95.9\%$\dag$ & -1021\% & -1138\% & -1132\% &       &   {}  &  {}  &  \\
        
        \cline{2-13}
        
        &  \textbf{+ $\mathcal{L_{\text{KL,uniform}}}$} & 2.8E-4 & 4.5E-4 & 8.2E-4 & 4.6E-4 & 1.125 & 0.993 & 0.996 & \multirow{2}[0]{*}{3} & \multirow{2}[0]{*}{5} & \multirow{2}[0]{*}{-} & \multirow{2}[0]{*}{-} \\
        
        &  \% drop & 99.6\%$\dag$ & 99.5\%$\dag$ & 98.2\%$\dag$ & 99.4\%$\dag$ & -14.5\% & -11.8\% & -11.5\% &       &   {}  &  {}  &  \\

    \hline
    \multirow{9}[0]{*}{\begin{tabular}[c]{@{}c@{}}\text{real-}\\\text{world}\end{tabular}} 
    
    & Base Model & 0.094 & 0.075 & 0.041 & 0.076 & 0.983 & 0.888 & 0.893 &       &       &       &  \\
    \cline{2-13}
    
    & + $\mathcal{L_{\text{KL,weighted\_adaptive}}}$ & 0.029 & 0.040 & 0.013 & 0.030 & 1.536 & 1.357 & 1.361 & \multirow{2}[0]{*}{6} & \multirow{2}[0]{*}{5} & \multirow{2}[0]{*}{0.80} & \multirow{2}[0]{*}{-} \\
        
    &  \% drop & 69.3\%$\dag$ & 46.2\%$\dag$ & 66.8\%$\dag$ & 59.9\%$\dag$ & -56.3\% & -52.8\% & -52.4\% &       &       &       &  \\

    \cline{2-13}
    &   \makecell[l]{+ $\mathcal{L_{\text{KL,weighted\_adaptive}}}$ - $\alpha$} & 0.032 & 0.041 & 0.019 & 0.033 & 1.148 & 1.000 & 1.005 & \multirow{2}[0]{*}{3} & \multirow{2}[0]{*}{5} & \multirow{2}[0]{*}{0.80} & \multirow{2}[0]{*}{-} \\
    
    &  \% drop & 66.6\%$\dag$ & 45.4\%$\dag$ & 52.9\%$\dag$ & 56.7\%$\dag$ & -16.8\% & -12.6\% & -12.5\% &       &       &       &  \\
    
    \cline{2-13}
    
    & + $\mathcal{L_{\text{KL,uniform}}}$ & 0.026 & 0.038 & 0.022 & 0.030 & 1.129 & 0.990 & 0.997 & \multirow{2}[0]{*}{5} & \multirow{2}[0]{*}{5} & \multirow{2}[0]{*}{-} & \multirow{2}[0]{*}{-} \\

    & \% drop & 72.7\%$\dag$ & 49.7\%$\dag$ & 45.1\%$\dag$ & 60.6\%$\dag$ & -14.9\% & -11.4\% & -11.6\% &       &       &       &  \\
    
    \cline{2-13}
    
    & \textbf{\makecell[l]{+ $\mathcal{L_{\text{KL,weighted\_adaptive}}}$ + $\mathcal{L}_{\text{MLM}}$}}
    
    & 0.027 & 0.032 & 0.016 & 0.027 & 1.256 & 1.080 & 1.084 & \multirow{2}[0]{*}{5} & \multirow{2}[0]{*}{5} & \multirow{2}[0]{*}{0.80} & \multirow{2}[0]{*}{0.1} \\

    & \% drop & 71.6\%$\dag$ & 57.0\%$\dag$ & 60.2\%$\dag$ & 64.6\%$\dag$ & -27.8\% & -21.6\% & -21.4\% &       &       &       &  \\

      \hline
    
    \end{tabular}
    {\captionsetup{skip=2pt}
    \caption{Bias mitigation and language modeling performance (\textbf{BERT-large}). See Table \ref{tab:result-distilbert} for cell values and notation details.}
  \label{tab:result-bert-large}
  }
  \vspace{-7mm}
\end{table*}
}
\raggedbottom

\subsubsection{Equal distribution}

\noindent Across all MLMs, applying uniform KL loss results in a consistent and substantial bias reduction. In all cases, bias was almost completely removed, with reduction exceeding 98\%, shifting the MLM's predicted gender distribution for each profession to a negligible deviation from the ideal 50\%-50\% male-female distribution.
This mitigation was observed consistently across all three profession categories, and ALL (all profession categories).
All the reductions are statistically significant (95\% confidence level), as determined by \textit{independent t-tests}.

Evaluating MLM loss on external corpora before and after bias mitigation, we find only small degradation, indicating that the language modeling capabilities were well preserved. Specifically, MLM loss increased by 2.3\% to 14.5\% (GAP: 4.7–14.5\%, WikiText-103-dev: 2.3–11.5\%, WikiText-103-test: 2.4–11.8\%).
For DistilBERT and BERT-base, degradation was minor ($<$ 8\%), while BERT-large showed slightly higher degradation (around 11–14\%). Additionally, across MLMs, GLUE scores (Appendix Table \ref{tab:mlm-glue-results} `debiased for equal' rows) remain consistent before and after debiasing, indicating preserved language modeling capabilities.
Overall, the results demonstrate that bias mitigation through uniform KL loss achieves near-complete bias removal with minimal compromise to language modeling performance.

\subsubsubsection{Baseline comparison}
\noindent \textbf{AttenD:} Results are presented in the rows ``AttenD'' (Tables \ref{tab:result-distilbert} - \ref{tab:result-bert-large}).
Our bias mitigation approach performs comparably to AttenD, achieving near-complete bias mitigation across all MLMs, with slightly better results on BERT-large.
However, our method preserves language modeling capability, whereas AttenD suffers a drastic MLM loss increase (over 1000\%), despite maintaining downstream GLUE performance (their Table 5). Our debiased model also preserves GLUE performance, so while bias mitigation is similar, the difference in MLM loss is stark.

\vspace{0.3em}
\noindent \textbf{CDS:} We do not re-run \citet{bartl-etal-2020-unmasking} in our setting as their debiasing relies on fine-tuning with the GAP corpus. In contrast, our method and AttenD operate directly on probe sentences derived from templates using profession and gendered words, making CDS results not directly comparable. For relative context, we report their results (Appendix Table \ref{tab:bartl-result}).
They measured bias via association score differences, which capture disparities in the model’s association between gender and profession, while we use distributional divergence. Despite different metrics, relative comparisons are insightful.

\citet{bartl-etal-2020-unmasking} reported $>$ 90\% mitigation for \DPM, 58\% for \DPF, and 68\% for \DPB. In contrast, our method achieves $>$ 98\% mitigation across all categories (Tables \ref{tab:result-distilbert}-\ref{tab:result-bert-large}), showing consistent, robust, and near-complete mitigation for equality across MLMs, underscoring effectiveness.

Moreover, our approach supports the alignment of model attribute–target distribution to a desired distribution – extending beyond conventional emphasis on equity in prior work.

\subsubsection{Real world distribution}

\noindent We first present the results using the weighted adapted KL loss, which constitutes our main bias mitigation method. We then evaluate the effect of adding a secondary MLM loss to preserve language modeling capability, yielding the best trade-off between fairness and language modeling performance. 
Finally, we conduct ablation studies by independently removing adaptive loss scaling and weighting (i.e., using uniform KL loss across professions) to assess the importance of these components.

\noindent \textbf{Weighted adaptive KL loss:} The rows ``+ $\mathcal{L_{\text{KL,weighted\_adaptive}}}$'' indicates the results of using weighted adaptive KL loss to reduce bias.
Bias reductions are significant with one exception: in BERT-base, for \DPF, the divergence slightly increased by 0.3\%. However, the initial bias was relatively low (0.04).
Bias reduction for \DPM ranged from 69\%-77\%, for \DPF from 46\%-57\%, for \DPB from 28\%-74\%, and for ALL from 59\%-67\%. 
These results demonstrate the effectiveness of the method across all profession categories.
However, the improvements came with a trade-off in degradation in MLM language modeling performance---MLM loss increased by 12\%-56\%. Notably, BERT-large exhibited the largest degradation (more than 50\%).

\noindent \textbf{Ablation: effect of adding adaptive loss scaling ($\alpha$):} Adaptive loss scaling controls the update magnitude. We assign a smaller $\alpha$ = $1\text{e-}6$ to groups with larger KL divergence (larger updates) and a larger $\alpha = 1\text{e-}5$ to groups with smaller KL divergence (smaller updates). Groups are defined using validation KL means (Appendix Table \ref{tab:validation-before-debaising}) of ALL: categories above the mean received a smaller $\alpha$, and those below received a larger $\alpha$. In all three MLMs, \DPM formed a larger update group, while \DPF and \DPB formed smaller update groups.

We compare the effect of removing adaptive loss scaling (``+ $\mathcal{L_{\text{KL,weighted\_adaptive}}}$ - $\alpha$'') with ``+ $\mathcal{L_{\text{KL,weighted\_adaptive}}}$''.
Removing scaling generally degrades bias mitigation.
In BERT-large, removal worsens bias mitigation performance across all categories, though language modeling performance improves significantly (about 40\% points). In BERT-base and DistilBERT, removing $\alpha$ slightly reduces bias mitigation for one of three categories and consistently across ALL slightly, alongside moderate degradation in MLM performance (1.3-8.5\% points). Overall, adaptive loss scaling yields slight but consistent improvement in bias mitigation, with trade-offs in modeling performance.

\noindent \textbf{Ablation: effect of weighted adaptive loss:} 
We now compare weighted adaptive loss
($\mathcal{L_{\text{KL,weighted\_adaptive}}}$) with non-adaptive uniform KL loss
($\mathcal{L_{\text{KL,uniform}}}$).

Using uniform loss, impacts bias mitigation and language modeling with trade-offs across models. In DistilBERT, uniform weighting yields a very small improvement for \DPM, but leads to a larger drop for \DPF (13.8\% points), and a moderate drop for \DPB (7.1\% points), along with a slight overall drop in ALL. However, MLM loss improves moderately (10.4\%-13.2\% points).
Notably, \DPM has a larger KL divergence initially than \DPF and \DPB. Uniform loss emphasizes the group with a larger KL, contributing more to the overall loss and improving bias mitigation for the dominant group, but causing a noticeable drop for non-dominant ones.

This pattern persists in BERT-base and BERT-large. In BERT-base, there is a very small drop for \DPM, while \DPF and \DPB experience moderate drops of 10.6\% points and 7.9\% points, respectively. Here, too, language modeling performance improves (7.3\%-10.5\% points), reflecting better preservation of MLM capability with uniform weighting.
In BERT-large, uniform loss yields small improvements for both \DPM and \DPF but a noticeable degradation for \DPB (21.7\% points). Importantly, MLM loss improves substantially (about 41\% points). 

\noindent \textbf{Weighted adaptive KL loss + MLM loss:} 
Weighted adaptive loss improves bias mitigation but reduces MLM performance. To address this trade-off, we add a secondary MLM loss to preserve language modeling ability while minimizing deviation from the target distribution.
Results are in the last rows of Tables \ref{tab:result-distilbert}-\ref{tab:result-bert-large}. We compare with the row ``+ $\mathcal{L_{\text{KL,weighted\_adaptive}}}$'' (without MLM loss).

In DistilBERT, there is a modest reduction in bias mitigation (6.6-9.2\% points) for \DPM and ALL, while stable or improved results for the rest. This fairness trade-off is accompanied by a substantial gain in language modeling, with MLM losses improving by 10.6-13.2\% points across corpora.
In BERT-base, bias mitigation performance remains largely stable or slightly improves (2\%-6\% points) after introducing MLM loss, while simultaneously achieving a 5-7\% points improvement in MLM loss across both corpora. BERT-large exhibits the most favorable outcome: bias mitigation performance improves across profession categories, alongside a substantial improvement in MLM loss (reduction of 28-31\% points). Downstream GLUE evaluation is maintained across all three MLMs, indicating preserved language modeling capabilities (Appendix Table \ref{tab:mlm-glue-results} `debiased for real-world' rows).
Overall, adding the MLM loss consistently improves language modeling performance (measured with MLM loss and maintained for GLUE evaluation), while modestly affecting bias mitigation performance (preserved or even improved slightly).

\noindent \textbf{Summary:} Weighted adaptive loss improves bias mitigation but trades off language modeling performance. Adding MLM loss reduces these trade-offs, maintaining (or only slightly reducing) bias mitigation while improving language modeling. Overall, weighted adaptive loss with MLM loss reduces MLM distribution deviation from the real-world target while preserving modeling performance.

\subsection{Results for debiasing ALM}

\noindent 
We present pre-debiasing results across Llama and Qwen Instruct models (3B - 72B), to examine how well ALM output distribution aligns with equality versus real-world target distribution across model sizes (Appendix Tables \ref{tab:bias-detection-results-alm-equal-dist} - \ref{tab:bias-detection-results-alm-real-world-dist}). KL divergence is near zero under the \textit{equal} distribution, indicating no bias, but notable bias is observed under \textit{real-world} distribution, with similar trends across model sizes and model type. Detailed discussions are provided in Appendix \ref{pre-debiasing-results-alms}. Now we will focus on debiasing results for two Llama instruct models (3.2-3B and 3.1-8B). Results are presented in Table  \ref{tab:result-llama-3.2-3B}.

{
\setlength{\tabcolsep}{3pt}
\renewcommand{\arraystretch}{1.1}

\begin{table*}[htb!]
    \centering
    \tiny
    \captionsetup{font=footnotesize}
    \begin{tabular}{lll|cccc|ccc|cccc}
    \hline
    \multirow{3}[0]{*}{\makecell[l]{\textbf{LLM}}}
    
    &\multirow{3}[0]{*}{\makecell[l]{\textbf{Desired}\\\textbf{Distribution}}}

    & \multirow{3}[0]{*}{\textbf{Model}}
    
    & \multicolumn{4}{c|}{\textbf{Profession Category}} 
    
    & \multicolumn{3}{c|}{\textbf{Perplexity}} & \multirow{3}[0]{*}{\textbf{Epoch}} & \multicolumn{1}{c}{\multirow{3}[0]{*}{\textbf{$\mathcal{B}$}}} & \multicolumn{1}{c}{\multirow{3}[0]{*}{\textbf{$\beta$}}} & \multirow{3}[0]{*}{\textbf{$\gamma$}} \\
    
    \cline{3-10}
    {} & {}   &   {}  & \multicolumn{1}{p{3.41em}}{\textbf{$\text{DP}_{\text{male}}$\newline{}(KL)}} 
    & \multicolumn{1}{p{4em}}{\textbf{$\text{DP}_{\text{female}}$\newline{}(KL)}} 
    & \multicolumn{1}{p{4.5em}}{\textbf{$\text{DP}_{\text{balanced}}$\newline{}(KL)}} & \multicolumn{1}{p{2.6em}|}{\textbf{ALL\newline{}(KL)}} 
    & \multicolumn{1}{p{5em}}{\textbf{GAP\newline{}corpus}} 
    & \multicolumn{1}{p{6em}}{\textbf{WikiText-103\newline{}(test)}} & \multicolumn{1}{p{6em}|}{\textbf{WikiText-103\newline{}(dev)}} &  &       &  &  \\

    \hline
    \multirow{4}[0]{*}{Llama3.2-3B-Instruct} 
    
    & \multirow{1}[0]{*}{equal} 
    
    & Base Model & 4E-4 & 7E-4 & 1E-4 & 4E-4 & 30.6 & 16.9 & 17.1 &       &       &       & \\

    \cline{2-14}
    
    {} & \multirow{3}[0]{*}{\makecell{real-\\world}} 
    
        & Base Model & 0.199 & 0.108 & 0.001 & 0.123 & 30.6 & 16.9 & 17.1 &       &       &       &  \\
    
        \cline{3-14}
        
        {} & {} & + $\mathcal{L_{\text{KL,weighted\_adaptive}}}$ & 0.089 & 0.051 & 0.004 & 0.057 & 32.4 & 17.4 & 17.6 & \multirow{2}[0]{*}{24} & \multirow{2}[0]{*}{5} & \multirow{2}[0]{*}{0.6} & \multirow{2}[0]{*}{-} \\
    
       {} & {}  &  \% drop & 55.1\%$\dag$ & 52.7\%$\dag$ & -255\% & 53.8\%$\dag$ & -6.1\% & -3.0\% & -3.1\% &       &       &       &  \\

     \hline

     \multirow{4}[0]{*}{Llama3.1-8B-Instruct} 

     & \multirow{1}[0]{*}{equal} 
    
    & Base Model & 2E-3 & 4E-4 & 2E-4 & 8E-4 & 25.6 & 12.5 & 12.7 &       &       &       & \\
     
     \cline{2-14}
     
      {} &    \multirow{3}[0]{*}{\makecell{real-\\world}} 
    
        & Base Model & 0.181 & 0.114 & 0.001 & 0.118 & 25.6 & 12.5 & 12.7 &       &       &       &  \\
    
        \cline{3-14}
        
        {} & {} & + $\mathcal{L_{\text{KL,weighted\_adaptive}}}$ & 0.069 & 0.057 & 0.003 & 0.051 & 26.7 &  12.6 & 12.8 & \multirow{2}[0]{*}{11} & \multirow{2}[0]{*}{5} & \multirow{2}[0]{*}{0.6} & \multirow{2}[0]{*}{-} \\
    
       {} & {}  &  \% drop & 61.9\%$\dag$ & 49.5\%$\dag$ & -187.1\% & 56.7\%$\dag$ & -4.2\% & -1.3\% & -1.4\% &       &       &       &  \\

    \hline
    
    \end{tabular}
    {\captionsetup{skip=2pt}
    \caption{Bias mitigation and language modeling performance:  Perplexity, LLM Evaluation Harness (Appendix Table \ref{tab:LM-Evaluation-Harness-Results}) for \textbf{Llama3 Instruct (3.1-8B and 3.2-3B)}. $ \text{lora}_{r} = 64, \frac{\text{lora}_{\alpha}}{\text{lora}_{r}}=0.25$. See Table \ref{tab:result-distilbert} for explanation of cell values and notations.}
  \label{tab:result-llama-3.2-3B}}
  \vspace{-6mm}
\end{table*}
}
\raggedbottom

For \textit{equal} target distribution, the initial KL divergence is close to zero, indicating the absence of bias. However, for \textit{real-world} target distribution, we find notable bias. We will now focus on bias mitigation for \textit{real-world} as a desired distribution.

As a reminder, we use sentence loss as a proxy for measuring association, normalized to the gender-profession distribution. Since it also reflects ALM language modeling, we do not include additional language modeling preservation loss as in MLMs.
As observed in MLMs, weighted adaptive loss has better performance compared to one without adaptive weighting, so we will only focus on weighted adaptive loss to mitigate bias in Llama3.
As seen in Appendix Table \ref{tab:validation-before-debaising}, \DPM and \DPF show higher initial bias and thus receive larger adaptive loss scaling ($\alpha$) than \DPB.

Applying the weighted-adaptive KL loss significantly improves bias mitigation across most profession categories.
KL divergence drops by 50\% - 62\% across \DPM, \DPF, and ALL. For \DPB, although the KL divergence worsens by 255\%/187\%, the absolute value remains very small, indicating that the profession from this group is already closely aligned with the real-world distribution.
Bias reductions are about the same for both models, with the larger model showing slightly better reduction for \DPM and slightly less reduction for \DPF. Irrespective of model size, bias mitigation remains a challenge for all profession categories, although it improves for larger models.

Meanwhile, degradation in language modeling, measured using perplexity, remains minimal. Larger model better preserves performance (e.g., Llama3.1-8B-Instruct: 1.3\%, Llama3.2-3B-Instruct: 3\% degradation on Wikitext-103-test). QLoRA fine-tuning on 8B-Instruct converged in fewer epochs than LoRA on 3B-Instruct, consistent with more expressive models requiring less training. Downstream performance (Appendix Table \ref{tab:LM-Evaluation-Harness-Results}) across five benchmarks is maintained, showing language modeling ability is not compromised. Overall, weighted adaptive KL loss achieves successful bias mitigation while largely preserving language modeling performance.

\section{Related Works}

\noindent \textbf{Bias perspective:} 
Prior bias work emphasizes equality (often framed as disparate impact, demographic parity, equalized odds, or equal opportunity \cite{gallegos-etal-2024-bias,mehrabi2021survey}), targeting fine-grained associations between demographic attributes and contexts (e.g., gendered term–profession pairs).
These works often equalize association scores using static \cite{bolukbasi2016man} embeddings or contextual embeddings from template-based probing \cite{shi2024general,guo-etal-2022-auto,garimella-etal-2021-intelligent} and co-occurrence analysis  \cite{dhamala2021bold,bordia-bowman-2019-identifying}.
Literature enforces pairwise parity within a domain (e.g., profession), but often overlooks broader distributional alignment.

In contrast, we adopt a coarser, distributional approach---shifting gender ratios in professions to 50\%-50\%. Beyond promoting \textit{equality}, we consider \textit{real-world} alignment motivated by concerns about LLM hallucinations and the need for trustworthy fact-checking grounded in actual distributions.
Depending on application, we frame bias relative to a desired distribution---equal (for social equity) or real-world (for factual grounding).

\noindent \textbf{In-processing bias mitigation:} Common approaches include modifying the model architecture (e.g., adding debiasing layers) \cite{xu-etal-2025-biasedit,kumar-etal-2023-parameter,lauscher-etal-2021-sustainable-modular} or restricting training to certain parameters (e.g., attention head, adapter layers \cite{yang-etal-2025-bias,masoudian-etal-2024-effective,gaci-etal-2022-debiasing,attanasio-etal-2022-entropy}), to isolate and suppress biased representations while preserving general knowledge. 

Another strategy modifies the loss function to encode fairness during training \cite{xu-etal-2025-biasedit,gallegos-etal-2024-bias,zheng-etal-2023-click,yogarajan2023tackling,garimella-etal-2021-intelligent}. E.g., \citet{guo-etal-2022-auto} uses Jensen-Shannon Divergence to enforce uniformity in gender-conditioned output distribution. \citet{woo2023compensatory} introduce KL-based regularization to reduce gender bias in stereotypical sentence representations, while preserving linguistic integrity on non-stereotypical content.

We propose a weighted adaptive KL loss that aligns LLM gender–profession distributions to a target, dynamically updating based on varying gender dominance to enable balanced bias mitigation across professions.

\section{Conclusion}

We present a new perspective of bias through the lens of \textit{desired distribution}---either equal or
real-world. We proposed a method that adjusts LLMs' gender-profession distribution toward a desired distribution (our primary objective), applied to 3 MLMs and 2 ALMs. To achieve this, we use a weighted adaptive KL loss combined with a secondary MLM loss for MLMs (not required for ALMs). Bias is measured via the KL divergence of the model’s distribution from the desired distribution.
We demonstrate the advantage of using this loss through two ablation studies. Adding the MLM loss improves language modeling with minimal impact on bias mitigation for MLMs. For ALMs, the KL loss alone effectively reduces bias across profession categories with minimal drop in language modeling performance.
We evaluate across three profession categories---male-dominated, female-dominated, and balanced. Overall, we show that LLM output distributions can be effectively aligned with the desired distribution.

\section{Limitations}

Our bias analysis is limited to binary gender, reflecting the availability of real-world distribution data for binary gender categories only.
We specifically address gender bias mitigation through distribution alignment within the context of professions, utilizing gender-profession data from the U.S. Bureau of Labor Statistics. 
We also acknowledge that our analysis is limited to US-based distributions and requires inclusive real-world distributions for global applicability.
Further investigation is needed to extend this work to other demographic groups, such as race.

Our analysis is based on a limited set of 225 professions, suggesting that expanding to a broader range could yield additional insights. Similarly, we employed only six templates for bias mitigation. However, we balanced the selection of rare and common templates in the training and testing sets. 
Additionally, our analysis is limited to a template-based approach, which is a common approach used for bias mitigation in LLMs. Bias mitigation for open-ended generation introduces additional complexities, such as maintaining generation quality and ensuring contextual relevance. We acknowledge that for improved generalizability and comprehensiveness, it is important to evaluate in these settings reflecting real-world language use, where autoregressive LLMs are typically more suitable. However, given the scope of our work, we leave this exploration for future research. In addition, our analysis does not comprehensively evaluate out-of-distribution scenarios, such as template variation (e.g., altering the order of attributes and targets or using more diverse template structures), which could provide a stronger test of generalizability. We leave such exploration for future work.

Another limitation arises in the fine-tuning of masked language models (MLMs) using additional MLM loss: we relied on a small set of probe sentences derived from training templates to preserve language modeling capability. Using a larger external corpus during fine-tuning could better preserve the model’s language modeling capabilities.

Finally, our bias mitigation efforts were focused on two variants of the Llama3 autoregressive language models (ALM). Extending this exploration to additional ALMs remains an important direction for future research.

\section{Ethical Considerations}
We limit our analysis to binary gender due to the availability of real-world distribution data for binary gender only.

Our bias viewpoint—where we align the model’s output distribution with real-world data—can introduce or preserve existing social biases in LLMs. However, we motivate this choice by its positive impact on reducing hallucinations and improving fact-checking, especially in high-stakes domains where factual accuracy is critical. We emphasize that such bias introduction must not be misused to justify inequitable outcomes; rather, it should be applied with transparency and only when the application context prioritizes factual alignment. Importantly, real-world distributions often reflect underlying systemic inequalities, and their use as a bias mitigation target should be carefully justified and assessed on a case-by-case basis. 

Moreover, our reliance on U.S.-based distributions may reinforce a geographically constrained perspective, underscoring the necessity of incorporating non-U.S. contexts for international deployment.

Future work should also aim to include non-binary and underrepresented gender identities to ensure a more inclusive and comprehensive fairness evaluation.

\bibliography{custom}

\appendix

\section{Appendix}

\subsection{Templates}\label{templates}

\begin{table}[htb!]
    \centering
    \scriptsize
    \captionsetup{font=footnotesize}
    \begin{tabular}{@{}ll@{}}
         \hline
         \textbf{TID} & \textbf{Templates}\\
         \hline
         $\text{T}_1$ & {[}DET/PRONOUN{]} {[}attribute{]}  is {[}ARTICLE{]} {[}target{]}. \\
         
         $\text{T}_2$ & {[}DET/PRONOUN{]} {[}attribute{]}  works as {[}ARTICLE{]} {[}target{]}. \\
         
         $\text{T}_3$ & {[}DET/PRONOUN{]} {[}attribute{]} wants to become {[}ARTICLE{]} {[}target{]}.\\

        $\text{T}_4$ & {[}DET/PRONOUN{]} {[}attribute{]} applied for the position of {[}target{]}.\\

        $\text{T}_5$ & {[}DET/PRONOUN{]} {[}attribute{]}, the {[}target{]} had a good day at work.\\
        $\text{T}_6$ & {[}DET/PRONOUN{]} {[}attribute{]} started a career as {[}ARTICLE{]} {[}target{]}.\\
         \hline
    \end{tabular}
    \caption{{Templates. TID: template id,  attribute: gendered-word, target: profession, DET: this, PRONOUN: my}}
    \label{tab:gender-bias-in-profession-general}
\end{table}

\subsection{Attribute}

\begin{table}[h]
    \centering
    \small
    \begin{tabular}{ll}
    
    \hline
    
    \textbf{Male gendered words} & \textbf{Female  gendered words}\\
    
    \hline
    he, man, brother & she, woman, sister \\
    
    son, husband, boyfriend & daughter, wife, girlfriend \\

    father, uncle, dad & mother, aunt, mom \\

    grandpa, grandfather & grandma, grandmother \\
    
     \hline
    \end{tabular}
    \caption{Attributes: Gendered words. These gendered words are preceded by DET: \textit{this} for man/woman, no DET/PRONOUN for he/she, while for remaining, PRONOUN is \textit{my} in the templates in Table \ref{tab:gender-bias-in-profession-general}.}
    \label{tab:gendered-words}
\end{table}

\newpage
\subsection{Profession distribution}

\begin{table}[h]
    \footnotesize
    \centering
    \begin{tabular}{lcccc}
    \hline
    {}   & \textbf{Train} & \textbf{Valid} & \textbf{Test} & \textbf{Total} \\
    \hline
    \textbf{\DPM} & 59    & 13    & 18    & 90 \\
    \textbf{\DPF} & 58    & 14    & 18    & 90 \\
    \textbf{\DPB} & 29    & 7     & 9     & 45 \\
    \hline
    \textbf{ALL} & 146   & 34    & 45    & 225 \\
    \hline
    \end{tabular}
    \caption{Distribution of professions (target).}
    \label{tab:profession_dist}
\end{table}

\subsection{Model hyperparameters and selection}\label{model-configs}

\noindent Following \citet{zhou2025assessing}, who observe that seed 42 often yields better performance in machine learning experiments, we fine-tune the model using seed 42 and select optimal hyperparameters based on validation KL divergence loss across epochs. We use the AdamW optimizer with a weight decay of 0.01. Final results are reported as the average over five seed runs using the selected configuration (see Section \ref{results}).

We implement fine-tuning using PyTorch, HuggingFace Transformers, and the PEFT library, running on NVIDIA A100 (80 GB) and A40 (45 GB) GPUs. On average, fine-tuning took 4 hours for MLMs and around 9 hours for the ALM.

\noindent \textbf{Fine-tuning convergence criteria:} We track the KL divergence loss across all professions on the validation set and stop fine-tuning if it fails to improve from the previous best by at least a threshold (0.0001 for \textit{equal}, 0.001 for \textit{real-world}) for \textit{n} = 5 consecutive steps (patience). Note that adaptive weighting is not applied during validation. KL divergence is computed uniformly across professions as in standard bias detection.

\vspace{0.5em}
\noindent \textbf{Profession batch:} 
We explore batch sizes of 5 and 8 for training and use a fixed batch size of 3 for validation. Note that we ensure each batch contains professions from the same profession group. 

\vspace{0.5em}
\noindent \textbf{Learning rate:} We evaluate the performance using a learning rate of 2e-5\footnote{We did preliminary analysis using a small subset of professions adapted from \citet{bartl-etal-2020-unmasking} using a learning rate of 1e-5 and 2e-5. We found 2e-5 converges earlier and performs better. For the small model DistilBERT, we also explored a smaller learning rate of 5e-6, but it degraded the performance in at least one of the profession groups and ALL (combined profession groups).} adapted from \cite{devlin-etal-2019-bert} for MLMs.
For Llama3.2-3B-Instruct, we evaluate using 2e-5 and 2e-4, adapted from \cite{hu2022lora}. For Llama3.1-8B-Instruct, we used the learning rate that performed best for Llama3.2-3B-Instruct.

\vspace{0.5em}
\noindent \textbf{Momentum weight for updating KL mean:} We explored momentum weights $\beta \in \{ 0.60, 0.80, 0.95\}$

\vspace{0.5em}
\noindent \textbf{Configurations for Llama3 fine-tuning using LoRA and QLoRA:} Following \citet{hu2022lora}, we fix attention dimension $\text{lora}_{r}$ = 64, which defines the rank of the low-rank adaptation and controls the number of trainable parameters. 
We vary the scaling factor $\text{lora}_{\alpha} \in \{16, 32,64\}$, where the ratio $\frac{\text{lora}_{\alpha}}{\text{lora}_{r} }\in \{0.25, 0.50, 1\}$ controls the strength of the LoRA update. A LoRA dropout of 0.2 is applied for regularization.

We apply LoRA to the projection layers in the model, including query (q\_proj), key (k\_proj), value (v\_proj), and output (o\_proj) projections in the self-attention mechanism, as well as the gate (gate\_proj), up (up\_proj), and down (down\_proj) projections in the MLP components. This setup allows LoRA to adapt both the attention and feedforward pathways, consistent with configurations shown to be effective in prior work on Llama-based instruction fine-tuning \cite{ibrahim-2024-cufe-stanceeval2024,pontes-etal-2024-l3itc,wang-etal-2025-pmss}. 

For models $\ge$7B parameters, such as Llama3.1–8B-Instruct, we apply QLoRA with 4-bit NF4 quantization (including double quantization and float16 compute) for both inference and bias mitigation. The same quantization configuration is used during pre-debiasing evaluation and QLoRA fine-tuning to ensure consistency. In contrast, for the smaller Llama3.2–3B-Instruct model, we use full-precision LoRA fine-tuning without quantization. Both models are fine-tuned for bias mitigation with identical hyperparameters.

\vspace{0.5em}
\noindent \textbf{Mean vs sum of KL divergence across gender:}
We observe that the MLM loss for male and female for a given profession varies, and to provide a balanced combined effect while computing the loss, we take the mean instead of the sum of individual divergence to get the total divergence for a profession. 

Our preliminary analysis using 60 professions (adapted from \citet{bartl-etal-2020-unmasking}; U.S. Bureau of Labor Statistics 2019) supports this. Irrespective of the \textit{mean} or \textit{sum} of KL divergence between the MLM-predicted and desired male-female distributions, convergence was similar for both equal and real-world distributions. However, 50\% of KL loss with \textit{sum} is achieved in the same epoch as \textit{mean}, i.e., to achieve similar performance, the sum requires further tuning. So, taking the mean divergence is optimal. 

\noindent \textbf{Selection of hyperparameter based on validation performance:} We use the fine-tuned model (obtained using seed 42) to evaluate performance on the validation set, using the corresponding validation templates and professions. 
We select the configuration that achieves high overall performance across all profession groups, as well as ALL (combined all three profession categories) with consistent (less spread) performance across three profession groups. We will discuss the approach next. 

We first compute the mean/standard deviation ($\frac{\mu}{\sigma}$), denoted as $R$, of performance improvement (relative to the Base model) across male-dominated, female-dominated, and balanced groups, aiming for a high mean with low variability, indicating strong and stable results. 
Runs are ranked by $R$ in descending order, and we select the top two whose ALL improvement exceeds the average ALL (allowing a small offset to include runs within 1 point of the mean). The final selection is based on the higher median improvement across the three profession groups, ensuring robustness to outliers and consistently strong improvement in at least half the groups.

\subsection{Common vs Rare templates}\label{common-rare-tempalte-dist}

\begin{table}[htb!]
  \centering
  \footnotesize
    \begin{tabular}{@{}cccc@{}}
    \hline
     {}   & \multicolumn{3}{c}{\textbf{\% sentences with ppl $<$ 15}} \\
     \cline{2-4}
    \multicolumn{1}{l}{\textbf{TID}} & \multicolumn{1}{l}{\textbf{DistilBERT}} & \multicolumn{1}{l}{\textbf{BERT-base}} & \multicolumn{1}{l}{\textbf{BERT-large}} \\
    \hline
    T1    & 41\%  & 48\%  & 46\% \\
    T2    & 63\%  & 70\%  & 71\% \\
    T3    & 77\%  & 74\%  & 80\% \\
    T4    & 73\%  & 86\%  & 87\% \\
    T5    & 85\%  & 98\%  & 99\% \\
    T6    & 40\%  & 46\%  & 39\% \\
    \hline
    \end{tabular}%
  \caption{Percentage of sentences with pseudo-perplexity (ppl) below 15}
  \label{tab:ppl_dist}%
\end{table}

\subsection{Bias on Validation set}

{
\setlength{\tabcolsep}{4pt}
\renewcommand{\arraystretch}{1.2} 
\begin{table}[htb!]
  \centering
  \tiny
    \begin{tabular}{l|Hcc|cc|cc|cc}
    \hline
         {} & {} & \multicolumn{8}{c}
          {\textbf{Profession category}} \\

    \cline{2-10}
        
    \textbf{Model} &
    \textbf{Desired Dist} & 
    \multicolumn{2}{c|}{\textbf{\DPM}} & 
    \multicolumn{2}{c|}{\textbf{\DPF}} & 
    \multicolumn{2}{c|}{\textbf{\DPB}} &
    \multicolumn{2}{c}{\textbf{ALL}} \\

    \cline{2-10}

    {} & {} & $\mu_{\text{KL}}$ & $\sigma^{2}_{\text{KL}}$  & $\mu_{\text{KL}}$ & $\sigma^{2}_{\text{KL}}$ & $\mu_{\text{KL}}$ & $\sigma^{2}_{\text{KL}}$ & $\mu_{\text{KL}}$ & $\sigma^{2}_{\text{KL}}$  \\

    \hline
    \multirow{1}[0]{*}{DistilBERT} 
          
    & real world & 0.166 & 0.019 & 0.036 & 0.003 & 0.025 & 3E-04 & 0.083 & 0.013 \\

    \multirow{1}[0]{*}{BERT-base} 
    
    & real world & 0.232 & 0.087 & 0.038 & 0.001 & 0.085 & 0.007 & 0.122 & 0.042 \\
    
    \multirow{1}[0]{*}{BERT-large} 
    
  & real world & 0.131 & 0.015 & 0.033 & 0.002 & 0.021 & 0.001 & 0.068 & 0.009 \\

    \multirow{1}[0]{*}{Llama3.2-3B} 
          
    & real world & 0.184 & 0.006 & 0.099 & 0.003 & 0.002 & 4E-06 & 0.112 & 0.008 \\

    \multirow{1}[0]{*}{Llama3.1-8B}

    & real world & 0.172 & 0.005 & 0.102 & 0.003 & 0.002 & 8E-06 & 0.108 & 0.007 \\

    \hline
    \end{tabular}
    \caption{Validation performance (KL mean: $\mu_{\text{KL}}$ and KL variance: $\sigma^{2}_{\text{KL}}$) across LLMs for \textbf{real world} as desired distribution (Before Debiasing)}

  \label{tab:validation-before-debaising}
\end{table}
}

\newpage
\subsection{\citet{bartl-etal-2020-unmasking} bias mitigation result}

{
\setlength{\tabcolsep}{2pt}
\renewcommand{\arraystretch}{1.7}

\begin{table}[htbp]
  \centering
  \tiny
  
    \begin{tabular}{lccccccc}

    \hline

    \makecell[l]{\textbf{Profession}\\\textbf{Category}} 
    & \makecell[l]{\textbf{Gender}} 
    & \makecell[l]{\textbf{Pre}\\\textbf{Mean}} 
    & \makecell[l]{\textbf{Post}\\\textbf{Mean}} 
    & \textbf{\makecell[l]{$|\text{m-f}|_{\text{Pre}}$} }
    & \textbf{\makecell[l]{$|\text{m-f}|_{\text{Post}}$} }
    & \textbf{\makecell[l]{$\Delta |\text{m-f}|_{\text{Post-Pre}}$}} 
    & \makecell[l]{\textbf{\% Bias}\\\textbf{Reduction}} \\

    \hline

    \multirow{2}[0]{*}{\DPB} & f     & -0.35 & 0.20   & \multirow{2}[0]{*}{0.40} & \multirow{2}[0]{*}{0.13} & \multirow{2}[0]{*}{0.27} & \multirow{2}[0]{*}{67.5\%} \\
    
    & m     & 0.05  & 0.07  &       &       &       &  \\
    
    \hline
    
    \multirow{2}[0]{*}{\DPF} & f     & 0.50   & 0.36  & \multirow{2}[0]{*}{1.18} & \multirow{2}[0]{*}{0.50} & \multirow{2}[0]{*}{0.68} & \multirow{2}[0]{*}{57.6\%} \\
          & m     & -0.68 & -0.14 &       &       &       &  \\

    \hline
    
    \multirow{2}[0]{*}{\DPM} & f     & -0.83 & 0.13  & \multirow{2}[0]{*}{0.99} & \multirow{2}[0]{*}{0.08} & \multirow{2}[0]{*}{0.91} & \multirow{2}[0]{*}{91.9\%} \\
          & m     & 0.16  & 0.21  &       &       &       &  \\
    \hline
          
    \end{tabular}
    \caption{Results are directly adapted from Table 4 in \citet{bartl-etal-2020-unmasking}. \textit{Pre} refers to the association scores between gender and profession before debiasing, and \textit{Post} refers to the scores after debiasing, based on the templates used in their paper.}
  \label{tab:bartl-result}
\end{table}
}

\subsection{Bias detection across ALMs}\label{pre-debiasing-results-alms}

{
\setlength{\tabcolsep}{3pt}
\renewcommand{\arraystretch}{1.5}

\begin{table*}[htb!]
    \centering
    \scriptsize
    \captionsetup{font=footnotesize}
    \begin{tabular}{l|cccc|ccc}
    \hline
    \multirow{3}[0]{*}{\makecell[l]{\textbf{LLM}}}
    
    & \multicolumn{4}{c|}{\textbf{Profession Category}} 
    
    & \multicolumn{3}{c}{\textbf{Perplexity}} \\
    
    \cline{2-8}
    
    {}   & \multicolumn{1}{p{3.41em}}{\textbf{$\text{DP}_{\text{male}}$\newline{}(KL)}} 
    
    & \multicolumn{1}{p{4em}}{\textbf{$\text{DP}_{\text{female}}$\newline{}(KL)}} 
    & \multicolumn{1}{p{4.5em}}{\textbf{$\text{DP}_{\text{balanced}}$\newline{}(KL)}} & \multicolumn{1}{p{2.6em}|}{\textbf{ALL\newline{}(KL)}} 
    & \multicolumn{1}{p{5em}}{\textbf{GAP\newline{}corpus}} 
    & \multicolumn{1}{p{6em}}{\textbf{WikiText-103\newline{}(test)}} & \multicolumn{1}{p{6em}}{\textbf{WikiText-103\newline{}(dev)}} \\

    \hline
    Llama3.2-3B-Instruct & 4E-4 & 7E-4  & 1E-4 & 4E-4 & 30.6 & 16.9 & 17.1 \\
    Llama3.1-8B-Instruct & 2E-3 & 4E-4 & 2E-4 & 8E-4 & 25.6 & 12.5 & 12.7 \\
    Llama3.3-70B-Instruct & 1E-3 & 6E-4 & 4E-4 & 9E-4 & 21.0 & 8.3 & 8.9 \\
    Qwen-2.5-7B-Instruct & 4E-3 & 5E-4 & 1E-3 & 2E-3 & 26.8 & 13.3 & 14.0 \\
    Qwen-2.5-72B-Instruct & 3E-3 & 5E-4 & 6E-4 & 2E-3 & 22.4 & 8.6 & 9.8 \\
    \hline
    
    \end{tabular}
    \caption{Pre-debiasing results across ALMs (\textit{equal} distribution) on test set}
  \label{tab:bias-detection-results-alm-equal-dist}
\end{table*}
}
\raggedbottom

{
\setlength{\tabcolsep}{3pt}
\renewcommand{\arraystretch}{1.5}

\begin{table*}[htb!]
    \centering
    \scriptsize
    \captionsetup{font=footnotesize}
    \begin{tabular}{l|cccc|ccc}
    \hline
    \multirow{3}[0]{*}{\makecell[l]{\textbf{LLM}}}
    
    & \multicolumn{4}{c|}{\textbf{Profession Category}} 
    
    & \multicolumn{3}{c}{\textbf{Perplexity}} \\
    
    \cline{2-8}
    
    {}   & \multicolumn{1}{p{3.41em}}{\textbf{$\text{DP}_{\text{male}}$\newline{}(KL)}} 
    
    & \multicolumn{1}{p{4em}}{\textbf{$\text{DP}_{\text{female}}$\newline{}(KL)}} 
    & \multicolumn{1}{p{4.5em}}{\textbf{$\text{DP}_{\text{balanced}}$\newline{}(KL)}} & \multicolumn{1}{p{2.6em}|}{\textbf{ALL\newline{}(KL)}} 
    & \multicolumn{1}{p{5em}}{\textbf{GAP\newline{}corpus}} 
    & \multicolumn{1}{p{6em}}{\textbf{WikiText-103\newline{}(test)}} & \multicolumn{1}{p{6em}}{\textbf{WikiText-103\newline{}(dev)}} \\

    \hline
    Llama3.2-3B-Instruct & 0.199 & 0.108 & 0.001 & 0.123 & 30.6 & 16.9 & 17.1 \\
    Llama3.1-8B-Instruct & 0.181 & 0.114 & 0.001 & 0.118 & 25.6 & 12.5 & 12.7 \\
    Llama3.3-70B-Instruct & 0.185 & 0.110 & 0.001 & 0.118 & 21.0 & 8.3 & 8.9 \\
    Qwen-2.5-7B-Instruct & 0.164 & 0.127 & 0.002 & 0.117 & 26.8 & 13.3 & 14.0 \\
    Qwen-2.5-72B-Instruct & 0.169 & 0.120 & 0.001 & 0.116 & 22.4 & 8.6 & 9.8 \\
    \hline
    
    \end{tabular}
    \caption{Pre-debiasing results across ALMs (\textit{real-world} distribution) on test set}
  \label{tab:bias-detection-results-alm-real-world-dist}
\end{table*}
}
\raggedbottom

\subsubsection{Equal distribution}
Table \ref{tab:bias-detection-results-alm-equal-dist} shows the pre-debiasing results using equality (50-50 gender-profession distribution) as the desired distribution across ALMs.

The initial KL divergence values remain close to zero for all ALMs irrespective of size. This indicates that there is no bias even in larger LLMs. In terms of language modeling capability, perplexity consistently decreases with model size, aligning with expectations—larger LLMs generally demonstrate stronger modeling performance. On average, perplexity drops (Llama 3B → 8B → 70B, Qwen 7B→72B) by approximately 16.9\% on the GAP corpus, 31.7\% on WikiText-103-test, and 28.6\% on WikiText-103-dev as model size increases. 

\subsubsection{Real world distribution}

Table \ref{tab:bias-detection-results-alm-real-world-dist} presents the pre-debiasing result using the real-world distribution as the desired distribution across ALMs.

We find similar levels of bias across model sizes. Consistently, male-dominated profession groups (\DPM) show higher bias than female-dominated ones (\DPF) and ALL profession categories, and there is no bias in the balanced (\DPB) category. Bias magnitude does not vary significantly with model size, thus, we observe a consistent trend with increasing size. The perplexity finding remains consistent with those discussed in the equal distribution section earlier.

\newpage
\subsection{Language modeling capability evaluation on downstream tasks}

{
\setlength{\tabcolsep}{2pt}
\renewcommand{\arraystretch}{1.7}

\begin{table*}[htbp]
  \centering
  \footnotesize
  \captionsetup{font=footnotesize}
    \begin{tabular}{llccccccccc}

    \hline
    MLM   & \multicolumn{1}{p{5.045em}}{\textbf{Desired}\newline{}\textbf{Distribution}} & \textbf{CoLA} & \textbf{SST-2} & \textbf{MRPC} & \textbf{STS-B} & \textbf{QQP} & \textbf{MNLI-(m/mm)} & \textbf{QNLI} & \textbf{RTE} & \textbf{Average} \\

    \hline

    \multirow{3}[0]{*}{DistilBERT} & Base Model & 0.49  & 0.91  & 0.90   & 0.86  & 0.87  & 0.82/0.82 & 0.89  & 0.60   & 0.79  \\
          \cline{2-11}
          & debiased for equal & 0.49  & 0.91  & 0.89  & 0.86  & 0.87  & 0.82/0.82 & 0.88  & 0.61  & 0.79 \\
          
          & debiased for real-world  & 0.50   & 0.91  & 0.89  & 0.86  & 0.87  & 0.82/0.82 & 0.89  & 0.60   & 0.80 \\

    \hline
    \multirow{3}[0]{*}{BERT-base} & Base Model & 0.56  & 0.93  & 0.88  & 0.88  & 0.88  & 0.84/0.85 & 0.91  & 0.62  & 0.82 \\
        \cline{2-11}
          & debiased for equal & 0.56  & 0.93  & 0.89  & 0.89  & 0.88  & 0.85/0.85 & 0.91  & 0.65  & 0.82 \\
          & debiased for real-world  & 0.56  & 0.93  & 0.89  & 0.89  & 0.88  & 0.85/0.85 & 0.91  & 0.64  & 0.82\\

    \hline
    \multirow{3}[0]{*}{BERT-large} & Base Model & 0.61  & 0.94  & 0.89  & 0.88  & 0.89  & 0.86/0.87 & 0.92  & 0.71  & 0.84  \\
        \cline{2-11}
          & debiased for equal & 0.59  & 0.94  & 0.90   & 0.90   & 0.88  & 0.86/0.86 & 0.92  & 0.71  & 0.84 \\
          & debiased for real-world  & 0.61  & 0.94  & 0.90   & 0.90   & 0.88  & 0.87/0.86 & 0.92  & 0.73  & 0.84 \\

    \hline
    
    \end{tabular}
    \caption{GLUE dev results across MLMs before and after mitigation (using weighted adaptive loss). Accuracy is reported for SST-2, MNLI, QNLI, and RTE; Spearman correlation for STS-B; Matthews correlation for CoLA; and F1 for MRPC and QQP. \textit{Base Model} refers to the pretrained model before debiasing, while \textit{debiased for equal} refers to models debiased by fine-tuning using non-adaptive KL loss, and \textit{debiased for real-world} refers to models debiased by fine-tuning using weighted adaptive loss combined with MLM loss. For each task, we report the average performance evaluated across five debiased models (obtained using random seeds 42, 52, 62, 72, and 82) for both equal and real-world target distributions.}
  \label{tab:mlm-glue-results}
\end{table*}
}

{
\setlength{\tabcolsep}{1.5pt}
\renewcommand{\arraystretch}{1.7}
\begin{table*}[htbp]
  \centering
  \scriptsize
  
    \begin{tabular}{cl|cccc|ccccccc|c}

    \hline
    \multirow{2}[0]{*}{\textbf{LLM}} & \multicolumn{1}{c|}{\multirow{2}[0]{*}{\textbf{Model}}} & \textbf{HellaSwag} & \multicolumn{1}{p{6.275em}}{\textbf{LAMBADA OpenAI}} 
    
    & \textbf{MMLU} & \textbf{TruthfulQA} & \textbf{CoLA} & \textbf{SST-2} & \textbf{MRPC} & \textbf{QQP} 
    
    & \textbf{MNLI-(m/mm)} 
    
    & \textbf{QNLI} & \textbf{RTE} & \multicolumn{1}{p{3.68em}}{\textbf{GLUE\newline{}Average}} \\

    \hline
    
    \multirow{2}[0]{*}{Llama3.1} & Base Model & 0.59  & 3.68/0.71 & 0.66  & 0.68/0.66/0.66 & 0.06  & 0.88  & 0.81  & 0.55  & 0.52/0.52 & 0.51  & 0.70  & 0.57 \\

    \cline{2-14}
          & + $\mathcal{L_{\text{KL,weighted\_adaptive}}}$ & 0.59  & 3.68/0.70 & 0.66  & 0.67/0.65/0.65 & 0.07  & 0.88  & 0.81  & 0.56  & 0.53/0.52 & 0.51  & 0.70  & 0.57 \\

    \hline
    
    \multirow{2}[0]{*}{Llama3.2} & Base Model & 0.52  & 3.67/0.71 & 0.60  & 0.66/0.64/0.64 & 0.03  & 0.83  & 0.83  & 0.56  & 0.52/0.52 & 0.54  & 0.74  & 0.58 \\

    \cline{2-14}
    
          & + $\mathcal{L_{\text{KL,weighted\_adaptive}}}$ & 0.53  & 3.67/0.71 & 0.60  & 0.67/0.65/0.65 & 0.03  & 0.83  & 0.82  & 0.56  & 0.53/0.52 & 0.53  & 0.73  & 0.58 \\

    \hline
    
    \end{tabular}
    \caption{LM Evaluation Harness for Llama3.2-3B-Instruct (Llama3.2) and Llama3.1-8B-Instruct (Llama3.1) before and after debiasing (using weighted adaptive KL loss). Metrics used — Accuracy: [HellSwag, MMLU, SST-2, MNLI, QNLI, RTE]; Perplexity/Accuracy: [LAMBADA OpenAI]; BLEU/ROUGE-1/ROUGE-L: [TruthfulQA\_Gen]; Matthews Correlation: [CoLA]; F1: [MRPC, QQP].}
  \label{tab:LM-Evaluation-Harness-Results}
\end{table*}
}

\label{sec:appendix}

\end{document}